\documentclass[sigconf]{acmart}
\usepackage{mathtools}
\usepackage{booktabs}
\usepackage{multirow}
\AtBeginDocument{%
  }

\setcopyright{cc}
\setcctype{by}
\copyrightyear{2026}
\acmYear{2026}
\acmDOI{10.1145/3831252.3834234}
\acmConference[ICCAD '26]{IEEE/ACM International Conference on Computer-Aided Design}{November 08--12, 2026}{San Jose, CA, USA}
\acmISBN{979-8-4007-2873-0/2026/11}
\acmBooktitle{IEEE/ACM International Conference on Computer-Aided Design (ICCAD '26), November 08--12, 2026, San Jose, CA, USA}

\begin{document}

\title[TARGet]{\textbf{TARGet}: \underline{\textbf{T}}opology-\underline{\textbf{A}}ware Fusion-based \underline{\textbf{R}}adio Frequency Circuit Functional Modeling using \underline{\textbf{G}}raph Neural N\underline{\textbf{et}}works}

\author{Soroosh~Noorzad}
\email{soroosh.noorzad@utah.edu}
\correspondingauthor
\orcid{0009-0004-4793-8327}
\affiliation{%
  \institution{University of Utah}
  \city{Salt Lake City}
  \state{Utah}
  \country{USA}
}

\author{Sebastian~Bodero}
\orcid{0009-0006-5714-9297}
\email{sebastian.bodero@utah.edu}
\affiliation{%
  \institution{University of Utah}
  \city{Salt Lake City}
  \state{Utah}
  \country{USA}
}

\author{Morteza~Fayazi}
\email{m.fayazi@utah.edu}
\orcid{0000-0002-5598-281X}
\affiliation{%
  \institution{University of Utah}
  \city{Salt Lake City}
  \state{Utah}
  \country{USA}
}
\renewcommand{\shortauthors}{Noorzad et al.}

\begin{abstract}
Automatic synthesis of analog and Radio Frequency (RF) circuits is an emerging area that requires an efficient circuit modeling method. In recent years, Machine Learning (ML) solutions have played a promising role in this regard. However, many existing ML approaches require separate training data for each circuit topology, even when a single circuit component is added or removed. In addition, they overlook circuit topology information, which limits their ability to capture complex component interactions. Furthermore, they rely on fully connected neural networks with flat feature representations, which require substantial amounts of training data. In this work, we propose an open-source topology-aware RF circuit modeling method, TARGet. Our model considers the circuit at two levels: sub-circuits and the overall circuit topology. At the sub-circuit level, TARGet leverages S-parameter representations to capture sub-circuit behavior rather than relying on individual circuit components, providing a reusable behavioral abstraction for RF building blocks. Moreover, TARGet explicitly incorporates circuit topology information into the model, enabling it to learn across multiple topologies. TARGet introduces a novel fusion-based architecture that integrates Graph Neural Networks (GNNs) and sub-circuit connectivity-aware neural networks to improve data efficiency. Experimental evaluation across multiple RF circuit topologies demonstrates that TARGet achieves sub-1\% prediction error while reducing the required training data by up to 35.5x compared to state-of-the-art (SOTA) approaches. Furthermore, TARGet achieves 9.7x higher prediction accuracy under a strict 1\% error threshold relative to SOTA models. A held-out matching-network evaluation further demonstrates zero-shot transfer to an unseen sub-circuit topology, where TARGet reduces NMAE by up to 45\%.
\end{abstract}




\maketitle

\section{Introduction}
Analog and Radio Frequency (RF) circuit design remains one of the most challenging tasks in Electronic Design Automation (EDA)~\cite{Fayazi2021Applications}. Unlike digital circuits, RF circuits exhibit strong nonlinearity and complex interactions among circuit components, making their performance highly sensitive to circuit topology and parameters. As a result, accurate circuit modeling plays a critical role in enabling automated synthesis, design space exploration, and optimization. In circuit modeling, the goal is to map circuit design parameters (\textit{e.g.}, component values, transistor sizes) to desired specifications (\textit{e.g.}, gain, power, bandwidth). On the other hand, optimization corresponds to the inverse problem of determining circuit design parameters that meet a given set of target specifications.

There are two classical approaches for automated synthesis of analog and RF circuits: simulation-based and model-based. In simulation-based methods, a circuit modeling tool is repeatedly used to evaluate circuits and generate new parameter candidates~\cite{Hassanpourghadi2021}, whereas in model-based approaches, a modeling tool is primarily leveraged to generate training data. Since both approaches depend heavily on a modeling tool, especially simulation-based ones, using conventional tools such as SPICE incurs a high computational cost. In addition, technology scaling has further exacerbated simulation runtime for both schematic and post-layout simulations~\cite{Fayazi2021Applications, Fayazi2023FuNToM}.

In recent years, Machine Learning (ML) techniques have emerged as a promising alternative to SPICE to accelerate RF circuit modeling by learning relationships between circuit parameters and performance metrics~\cite{AFACAN2021, zhao2021efficient, Liu2020, Hassanpourghadi2021, li2012efficient, Wang2016, fang2014bmf, Fukuda2017, Daems2003, McConaghy2005, Li2012, Alawieh2018, Abbineni2026MuaLLM}. Despite promising results, they have limited generalization across unseen topologies. This necessitates separate training data for each circuit topology, even for minor structural variations such as adding or removing a single component.

FuNToM~\cite{Fayazi2023FuNToM} has addressed modeling across multiple circuit topologies by leveraging an S-parameter-based modular representation method. FuNToM decomposes circuits into multiple sub-circuits and learns their behavior through a combination of sub-models and a global model. However, it does not explicitly incorporate the high-level circuit topology information. Consequently, the lack of topology awareness requires a separate model for each high-level circuit topology, significantly increasing data requirements and limiting the scalability of such approaches.

In addition, most existing techniques rely on Fully Connected Neural Networks (FCNNs) that operate on flat feature representations~\cite{Daems2003, McConaghy2005, Fukuda2017, Wolfe2003}. Such architectures are inherently mismatched with circuit structures, where component connectivity determines signal flow and overall behavior. By ignoring these structural dependencies, FCNNs fail to effectively capture topological information, leading to poor generalization and reduced data efficiency.

To address all the aforementioned challenges, we propose TARGet, an open-source\footnote{\url{https://github.com/medal-ece/TARGet}} topology-aware framework for RF circuit modeling. TARGet fuses different Neural Network (NN) models, such as Graph Neural Networks (GNNs) and modified versions of the Circuit Connectivity-Inspired Neural Network (CCI-NN)~\cite{Hassanpourghadi2021}, to achieve improved performance. The goals of TARGet are threefold: (a) reducing the required training data; (b) developing a unified model across different circuit topologies; and (c) enabling fast and accurate RF functional modeling.

To reduce the required training data, instead of relying on individual circuit components as model inputs, TARGet leverages S-parameter representations to capture sub-circuit behavior, providing a reusable behavioral description of each RF building block across circuit compositions~\cite{Fayazi2023FuNToM}. Moreover, it explicitly incorporates both high-level circuit topology and sub-circuit connectivity information as a dedicated representation. These abstractions enable learning from a unified dataset and model across multiple circuit topologies while supporting modular representations of sub-circuits.

TARGet introduces a fusion-based architecture that preserves circuit structure. Unlike FCNN-based models, which treat all input design parameters as a single flat vector, TARGet organizes inputs into separate sub-circuit groups. It further incorporates both high-level circuit topology and sub-circuit interconnectivity within a unified framework. The interconnectivity-aware part captures signal propagation and interactions among sub-circuits~\cite{Hassanpourghadi2021}. In parallel, a GNN encodes the high-level circuit topology by modeling the connections, values, and placements of circuit components and sub-circuits. By integrating these two complementary perspectives within a unified learning framework, TARGet further reduces the amount of training data required.

We evaluate TARGet across multiple RF circuit topologies, including Low-Noise Amplifiers (LNAs) and phase shifters. Experimental results demonstrate that TARGet achieves normalized error below \textbf{1\%}. Moreover, compared to state-of-the-art (SOTA) approaches, while maintaining the same error, TARGet reduces the required training data by up to \textbf{35.5x}. Furthermore, TARGet achieves \textbf{9.7x} higher prediction accuracy at a strict 1\% error threshold compared to SOTA models. To further evaluate topology-level generalization, we consider a stricter held-out setting in which one LNA matching-network sub-circuit topology is completely excluded from the training, validation, and test sets and reserved only for zero-shot evaluation. TARGet consistently outperforms all baselines on this split, reducing NMAE by up to \textbf{45.1\%}.

Our main contributions are summarized as follows:
\begin{itemize}
    \item Proposing an open-source topology-aware approach that enables unified RF circuit modeling across multiple topologies with normalized error below 1\%.
    \item Introducing a novel fusion-based technique that jointly integrates a GNN and a modified CCI-NN model.
    \item Reducing the required training data by up to 35.5x compared to the SOTA models.
    \item Achieving 9.7x higher prediction accuracy at a strict 1\% error threshold compared to SOTA models.
    \item Demonstrating zero-shot transfer to a held-out topology, where TARGet reduces NMAE by up to 45.1\% compared to the best baseline.
\end{itemize}

\section{Background \& Related Work}
\subsection{Problem Formulation}
Functional modeling aims to learn a mapping from circuit design parameters, $\mathbf{x}$, to a performance specification vector, $y$, \textit{i.e.},
\begin{gather}
\label{eq:model_goal}
\left\{
\begin{array}{l}
y = f(\mathbf{x})\\
y = \hat{f}(\mathbf{x}) + e,\\
\text{subject to: }
\text{minimize } e.\\
\end{array} \right.
\end{gather}
A learned model, $\hat{f}$, approximates this mapping with a modeling error of $e$.

Existing approaches for RF circuit modeling include posynomial regression, genetic programming, Bayesian Model Fusion (BMF), and NNs~\cite{Daems2003, McConaghy2005, Li2012, Fukuda2017}. These methods significantly reduce the need for expensive SPICE simulations by enabling fast prediction of circuit behavior. However, they require separate training data for each circuit topology and fail to generalize across structurally different circuits.

\subsection{Hierarchical Circuit Modeling}
To improve data efficiency and scalability, several works have incorporated circuit structural information into the modeling process. Hierarchical approaches such as Co-Learning Bayesian Model Fusion (CL-BMF) decompose circuits into multiple stages and learn lower-dimensional models~\cite{wang2015co, Alawieh2018}.
Similarly, CCI-NNs encode connectivity patterns within neural architectures to better reflect circuit behavior~\cite{Hassanpourghadi2021}. The CCI-NN approach incorporates the circuit's hierarchy into its modeling by using two types of connections between sub-circuits: sequential and direct paths. Other methods adopt multi-level modeling strategies, such as learning transistor-level behavior and using it for circuit-level prediction~\cite{zhao2021efficient}.

While these approaches partially exploit circuit structure, they rely on predefined decompositions or implicit representations of connectivity. Consequently, they do not explicitly model high-level circuit topology and remain limited in their ability to generalize across unseen circuit configurations.

\subsection{GNNs for Modeling}

\begin{figure*}[!t]
\centering
\includegraphics[width=\textwidth]{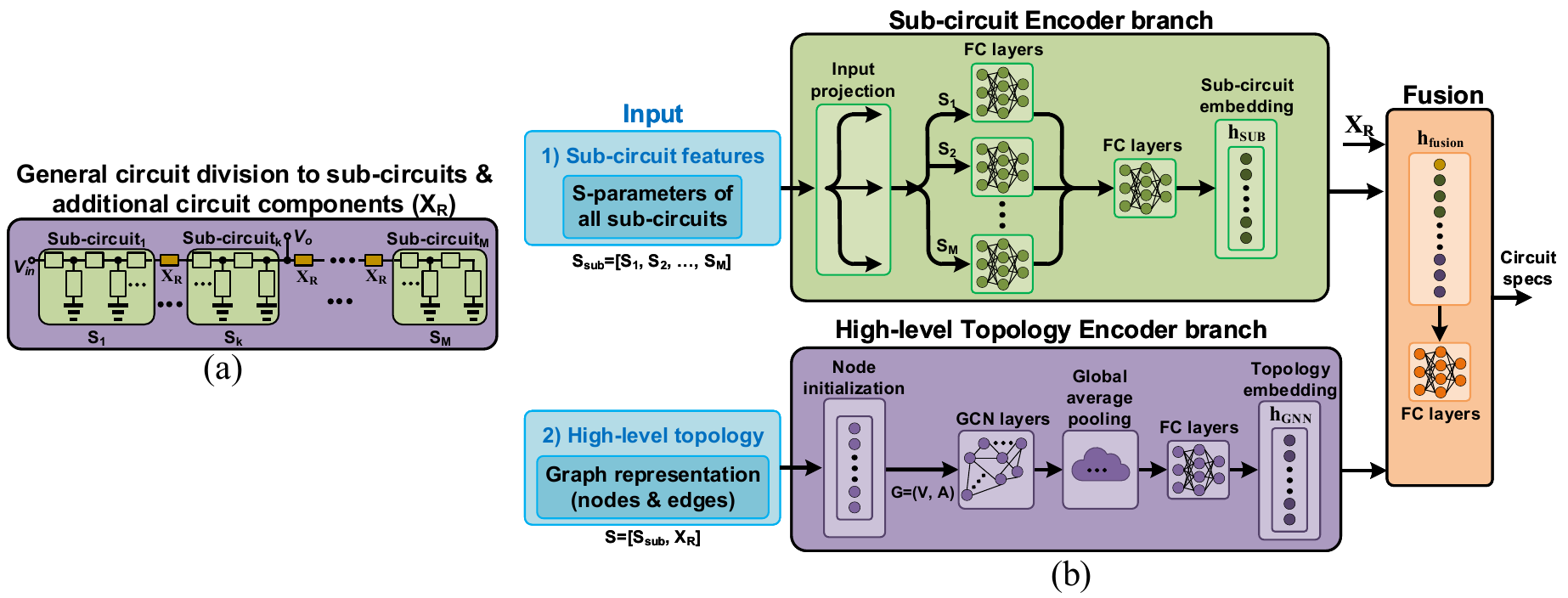}
\caption{Overall TARGet framework.}
\Description{Block diagram of TARGet. A circuit is divided into sub-circuits and auxiliary components. S-parameter features are processed by a sub-circuit encoder branch, while a graph representation of nodes and edges is processed by a high-level topology encoder branch. The two embeddings, together with auxiliary features, are concatenated in a fusion block and passed through fully connected layers to predict circuit specifications.}
\label{fig:framework}
\end{figure*}

GNNs provide a natural framework for modeling structured systems.
In circuit modeling, circuits are represented as graphs $G = (V, E)$, where nodes, $V$, correspond to circuit components or sub-circuits while edges, $E$, represent their connections~\cite{Lopera2021, Sanchez2022, zhang2019circuit}.
By aggregating information from neighboring nodes, GNNs learn representations that explicitly capture circuit topology.

Due to GNNs' ability to model relational structure, they have been applied to various problems in electronic design automation, including parasitic prediction~\cite{Ren2020, Liu2021parasitic}, transistor sizing~\cite{Wang2020gcnrl}, and performance modeling across multiple topologies~\cite{Wu2022Transfer}. However, existing GNN-based approaches for circuit modeling are often restricted to specific circuit types or require retraining for different topologies~\cite{Ren2020, Wang2020gcnrl, Wu2022Transfer, CircuitGNN2023, GraphOfCircuits2023}.
Moreover, purely graph-based models may not fully capture complex local interactions within circuit components~\cite{Lopera2021, Sanchez2022, CircuitGNN2023}. These limitations highlight the need for hybrid approaches that combine topology-aware modeling with sub-circuit connectivity-aware learning, enabling improved generalization and data efficiency across diverse RF circuit topologies.

\section{Proposed Approach}
\subsection{Framework Overview}
TARGet is designed to jointly model sub-circuit behavior and high-level topology within a unified learning architecture. As shown in Fig.~\ref{fig:framework}, the framework takes two types of inputs: The first corresponds to sub-circuit representations, which capture their electrical behavior. The second corresponds to high-level topology, which describes the interconnections among these blocks and other circuit components.

To learn from these two complementary input sources of information, TARGet adopts a dual-branch architecture. The first branch, \textit{Sub-circuit Encoder}, models interactions among sub-circuit representations. The second branch, \textit{Topology Encoder}, captures structural dependencies based on the high-level topology, which makes TARGet a topology-aware architecture. The outputs of these two branches are combined through a fusion mechanism to predict the circuit performance. This architecture enables TARGet to leverage both sub-circuit behavior and high-level topology within a unified framework, improving generalization across different circuit topologies.

\subsection{Input Representation} \label{subsec:input_rep}
To enable topology-aware modeling, TARGet adopts a modular representation of circuits, where each circuit is described in terms of sub-circuit-level abstractions rather than individual components. Here, modularity refers to representing circuit behavior using functional units that capture the input–output characteristics of groups of components. This representation allows functionally similar structures to share a common representation across different circuit implementations. Based on this formulation, each circuit is described using two complementary representations: (i) sub-circuit behavior encoded through S-parameters, and (ii) an explicit representation of circuit topology.

\subsubsection{\textbf{Sub-circuit Representation via S-parameters}} \label{subsubsec:subcir_rep}
A sub-circuit is defined as a group of interconnected electrical components treated as a single functional unit. Instead of using individual circuit component values, $\mathbf{x}^{(i)}$ (\textit{e.g.}, capacitor and inductor values), TARGet represents each sub-circuit using its corresponding S-parameters, $\mathbf{S}^{(i)}$. S-parameters provide a compact and expressive characterization of the electrical behavior of RF sub-circuits across frequency. Specifically, for the $i^{\text{th}}$ sub-circuit, the two-port scattering parameters are given by:

\begin{align}
\label{eq:s_param}
\mathbf{S}^{(i)}(f) =
\begin{bmatrix}
S_{11}^{(i)}(f) & \quad S_{12}^{(i)}(f) \\[8pt]
S_{21}^{(i)}(f) & \quad S_{22}^{(i)}(f)
\end{bmatrix},
\end{align}
where $f$ denotes the frequency. Due to the complex-valued nature of S-parameters, each entry is expressed in either magnitude-phase or Cartesian form. TARGet adopts the Cartesian representation, \textit{i.e.}:
\begin{align}
\label{eq:cartesian}
S_{JK}^{(i)} = \Re\left\{S_{JK}^{(i)}\right\} + j\, \Im\left\{S_{JK}^{(i)}\right\},
\end{align}
which is more suitable for neural network learning.

\subsubsection{\textbf{Topology Representation}} \label{subsubsec:topo_rep}
\begin{figure}[t]
\centering
\includegraphics[width=\columnwidth]{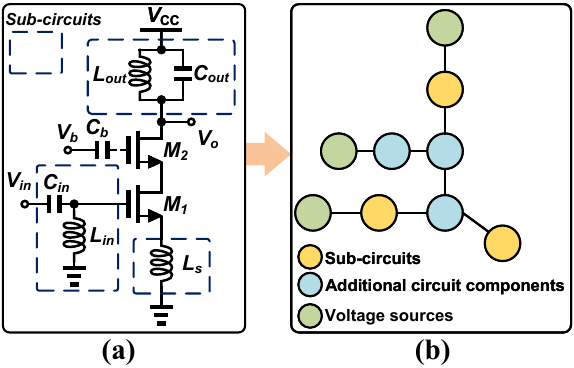}
\caption{An example of a circuit to graph decomposition in TARGet. (a) An RF circuit. (b) The corresponding graph.}
\Description{Two-panel example showing an RF circuit schematic on the left and its graph representation on the right. Dashed boxes mark sub-circuits in the schematic. The graph uses different node colors for sub-circuits, additional circuit components, and voltage sources, with edges representing electrical connections.}
\label{fig:circuit_to_graph_example}
\end{figure}

Each circuit is represented as a graph $G = (V, E)$, where nodes, $V$, correspond to circuit components or sub-circuits, and edges $E$ represent the electrical connections between them. An example of such a circuit-to-graph decomposition is shown in Fig.~\ref{fig:circuit_to_graph_example}. This graph-based representation enables the model to capture relationships across the entire circuit beyond sub-circuit representations.

Each node is represented by a vector consisting of three components: (i) a scalar indicating the node type, $\mathbf{NT}$, (ii) a feature vector encoding node-specific information, $\mathbf{NF}$ (\textit{e.g.}, S-parameters for sub-circuit nodes), and (iii) a one-hot identifier, $\mathbf{NID}_{oh}$. The node feature matrix $\mathbf{V}$ is formed by stacking node feature vectors $\mathbf{v}^{(i)}$, \textit{i.e.}:
\begin{align}
\mathbf{V} = [\mathbf{v}^{(1)}, \mathbf{v}^{(2)}, \ldots, \mathbf{v}^{(N)}]^{\top},
\end{align}
where $N$ is the number of nodes and 
\begin{align}
\label{eq:node_vector}
\mathbf{v}^{(i)} = 
\left[
\mathbf{NT}^{(i)},\;
\mathbf{NF}^{(i)},\;
\mathbf{NID}_{oh}^{(i)}
\right].
\end{align}

This matrix, together with the adjacency matrix describing connectivity, is used as input to the GNN layers.

\subsection{Sub-circuit Connectivity Modeling} \label{subsec:Sub-circuitConnectivityModelingViaCCI-NN}
While S-parameter representations provide a compact abstraction of sub-circuit behavior, directly processing them using conventional FCNNs remains suboptimal, as such models treat all inputs as a flat vector and ignore structured relationships among sub-circuits. This limits their ability to capture interaction patterns across different parts of the circuit~\cite{Hassanpourghadi2021}.

To address this limitation, TARGet employs a sub-circuit connectivity\nobreakdash-aware neural network, inspired by CCI-NN~\cite{Hassanpourghadi2021}. This model captures interactions among sub-circuits through grouped input processing. Unlike FCNNs, which jointly process all inputs, TARGet employs a CCI-based architecture that organizes inputs into sub-circuit-level groups and processes them using dedicated sub-models, enabling structured feature extraction.

\subsubsection{\textbf{Grouped Input Processing}} \label{subsubsec:GrInProc}
Let $\mathbf{S}^{(i)}$ denote the S-parameter representation of the $i^{\text{th}}$ sub-circuit, where $i = 1, \dots, M$ and $M$ is the number of sub-circuits. Each sub-model learns an intermediate representation:
\begin{align}
\mathbf{h}^{(i)} = \phi_{s}^{(i)}(\mathbf{S}^{(i)}),
\end{align}
where $\phi_{s}^{(i)}(\cdot)$ denotes a multi-layer perceptron for the $i^{\text{th}}$ sub-circuit.

\subsubsection{\textbf{Connectivity-Aware Feature Combination}} \label{subsubsec:ConAwCom}
To capture interactions among sub-circuits, TARGet combines intermediate representations obtained from individual sub-circuit models. Specifically, the learned representations $\mathbf{h}^{(i)}$ are combined through a structured aggregation of sub-circuit representations. The final sub-circuit embedding is defined as:
\begin{align}
\mathbf{h}_{\text{SUB}} = \psi_{\text{SUB}}\left([\mathbf{h}^{(1)}, \mathbf{h}^{(2)}, \dots, \mathbf{h}^{(M)}]\right),
\end{align}
where $\psi_{\text{SUB}}(\cdot)$ denotes a lightweight fully connected network. This design ensures that the sub-circuit encoder focuses on learning sub-circuit-level interaction patterns, while the topology encoder handles dependencies defined by the high-level topology.

\begin{table}[t]
\centering
\caption{Statistics of phase-shifter database.}
\label{tab:spec_stat_phase_shifter}
\begin{tabular}{lcccc}
\toprule
Spec & Min & Max & Average & SD\\
\midrule
Input return loss [dB] & -116.94 & 0 & -7.24 & 7.84\\
\midrule
Insertion loss [dB] & -153.5 & 0 & -5.46 & 8.9\\
\midrule
Output return loss [dB] & -116.94 & 0 & -7.24 & 7.84\\
\bottomrule
\end{tabular}
\end{table}

\begin{table}[t]
\centering
\caption{Statistics of LNA database.}
\label{tab:spec_stat_lna}
\begin{tabular}{lcccc}
\toprule
Spec & Min & Max & Average & SD\\
\midrule
Power gain (GP) [dB] & -226.7 & 51.95 & -12.08 & 16.68\\
\midrule
Available gain (GA) [dB] & -226.7 & 11.25 & -7.33 & 13.98\\
\midrule
Noise Figure (NF) [dB] & 2.41 & 3082.55 & 19.11 & 22.15\\		
\bottomrule
\end{tabular}
\end{table}

\begin{table}[t]
\centering
\footnotesize
\caption{Architectural details of TARGet and baseline models}
\label{tab:architecture_all}
\setlength{\tabcolsep}{4pt}
\begin{tabular*}{\linewidth}{@{\extracolsep{\fill}} l l p{3.5cm}}
\toprule
NN Model & Module & Configuration \\
\midrule
FCNN & Hidden layers & FC [128, 256, 128] \\
\midrule
\multirow{2}{*}{CCI-NN}
& Sub-ANNs & FC branches [128, 128, 128] \\
& Aggregation &  FC [512] \\
\midrule
\multirow{3}{*}{GNN}
& Graph encoder & GCN layers [64, 64, 64] \\
& Pooling & Global Avg Pool \\
& Output head & FC [64, 32] \\
\midrule
\multirow{6}{*}{TARGet}
& \multirow{2}{*}{Sub-circuit encoder} & FC branches [64, 64] \\
& & FC [128, 64] \\
\cmidrule(lr){2-3}
& \multirow{2}{*}{Topology encoder} & GCN layers [64, 64, 64] \\
& & Global Avg Pool \\
\cmidrule(lr){2-3}
& \multirow{2}{*}{Output head (Fusion)} & Concatenation \\
& & FC [64, 32] \\
\bottomrule
\end{tabular*}
\par\smallskip
{\footnotesize\raggedright \textit{Note:} FC stands for fully connected layers.\par}
\end{table}

\begin{table}[t]
\centering
\caption{Architectural comparison of TARGet and baseline models}
\begin{tabular}{l c c c c}
\toprule
Model & Input & Sub-circuit & Topology & Fusion \\
\midrule
FCNN & Flat & -- & -- & -- \\
CCI-NN & Grouped & \checkmark & -- & -- \\
GNN & Graph & -- & \checkmark & -- \\
TARGet & Hybrid & \checkmark & \checkmark & Late fusion \\
\bottomrule
\end{tabular}
\label{tab:model_comparison}
\end{table}

\begin{figure*}[t]
\centering
\includegraphics[width=\textwidth]{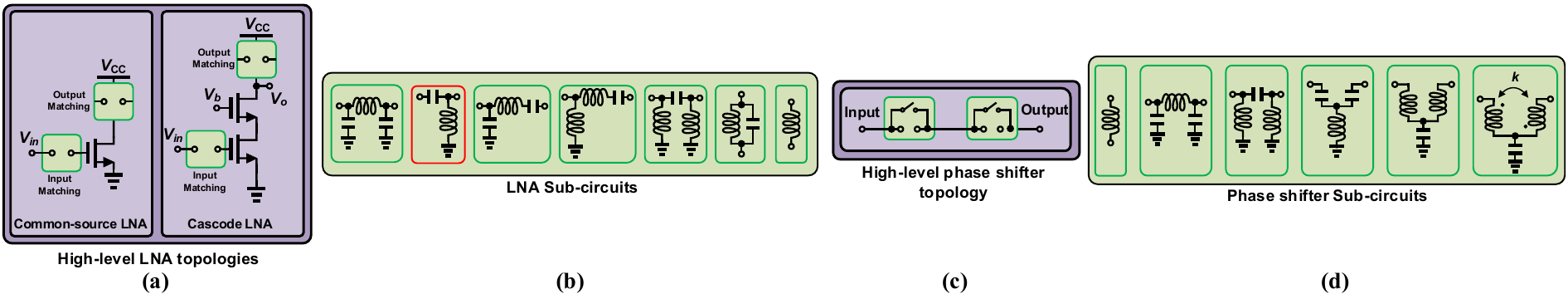}
\caption{Supported high-level topologies and sub-circuits. (a) LNA high-level topologies. (b)  LNA sub-circuits. (c) Phase shifter high-level topology. (d)  Phase shifter sub-circuits. For the zero-shot evaluation, one LNA sub-circuit topology is reserved as a held-out structure for zero-shot testing (red borders).}
\Description{Four-panel overview of the evaluated circuit families. The first panel shows common-source and cascode LNA high-level topologies with input and output matching blocks. The second panel shows the LNA matching-network sub-circuit library, with the held-out zero-shot topology marked by a red border. The third panel shows the high-level phase-shifter topology, and the fourth panel shows the phase-shifter sub-circuit library.}
\label{fig:database}
\end{figure*}

\subsection{High-level Topology Modeling via GNN} \label{subsec:High-levelTopologyModelingViaGNN}
While the sub-circuit encoder captures interactions among sub-circuits, it does not explicitly model the high-level topology of the circuit. To address this, TARGet incorporates a GNN to encode high-level topology.

The GNN learns node representations by iteratively aggregating information from neighboring nodes. Given a circuit graph $G = (V, E)$ defined in Section \ref{subsubsec:topo_rep}, let $\mathbf{V}$ denote the node feature matrix and $\mathbf{A}$ the corresponding adjacency matrix representing node connectivity. The initial node representations are given by $\mathbf{H}^{(0)} = \mathbf{V}$. At each layer $l$, the node representations are updated as:
\begin{align}
\mathbf{H}^{(l+1)} = \sigma\left(\tilde{\mathbf{A}}\, \mathbf{H}^{(l)}\, \mathbf{W}^{(l)} \right),
\end{align}
where $\mathbf{H}^{(l)}$ denotes node embeddings at layer $l$, the $i^{\text{th}}$ row of $\mathbf{H}^{(l)}$ is denoted as $\mathbf{h}_i^{(l)}$, $\mathbf{W}^{(l)}$ are learnable weights, $\sigma(\cdot)$ is a nonlinear activation function, and $\tilde{\mathbf{A}}$ is a normalized adjacency matrix that stabilizes feature aggregation across nodes.

Through this message-passing mechanism, each node progressively incorporates information from its neighbors, enabling the model to capture dependencies defined by the circuit topology. After multiple layers, node representations encode both sub-circuit features and higher-order (multi-hop) structural context, where nodes are influenced not only by their immediate neighbors but also by more distant components through successive aggregation steps.

To obtain a graph-level representation, the final node representations are aggregated using global average pooling:
\begin{align}
\mathbf{h}_{\text{GNN}} = \frac{1}{N} \sum_{i=1}^{N} \mathbf{h}_i^{(L)},
\end{align}
where $L$ is the number of GNN layers.

By explicitly encoding the high-level topology, the GNN enables TARGet to generalize across circuits with different topologies without requiring topology-specific models.

\subsection{Fusion Mechanism} \label{Fusion Mechanism}
TARGet integrates complementary representations learned by the sub-circuit encoder and topology encoder through a unified fusion mechanism. The sub-circuit branch captures structured interactions during feature extraction, while the topology branch encodes multi-hop dependencies induced by the circuit's high-level topology. Their combination enables joint reasoning over sub-circuit signal propagation and high-level topological relationships.

Let $\mathbf{h}_{\text{SUB}}$ denote the sub-circuit embedding produced by the sub-circuit encoder, and $\mathbf{h}_{\text{GNN}}$ denote the topology-aware graph embedding obtained from the topology encoder. These representations are concatenated to form the fused representation:
\begin{align}
\label{eq:fusion_concat_wo_xr}
\mathbf{h}_{\text{fusion}} = \left[ \mathbf{h}_{\text{SUB}}, \mathbf{h}_{\text{GNN}} \right].
\end{align}
Optional auxiliary features (\textit{e.g.}, circuit parameters that are not part of sub-circuit properties) can also be incorporated into the fusion stage when available. In this case, the representation becomes:
\begin{align}
\label{eq:fusion_concat_with_xr}
\mathbf{h}_{\text{fusion}} = \left[ \mathbf{h}_{\text{SUB}}, \mathbf{h}_{\text{GNN}}, \mathbf{X_R} \right].
\end{align}

The fused representation is passed through a lightweight multi-layer perceptron (MLP) to generate the final prediction:
\begin{align}
\hat{y} = f_{\text{head}}(\mathbf{h}_{\text{fusion}}),
\end{align}
where $f_{\text{head}}(\cdot)$ denotes the regression head.

This late-fusion design allows each branch to specialize in modeling distinct aspects of the circuit while maintaining a unified learning framework. By separating sub-circuit connectivity modeling from high-level topology reasoning, TARGet improves data efficiency and generalization across circuit topologies. Unlike early fusion strategies that combine heterogeneous features at the input level, the proposed approach preserves the inductive biases of each branch and integrates them only at the decision stage.

\subsection{Training Objective}
TARGet is trained in a supervised manner to learn the mapping between circuit representations and performance specifications. Given training samples, the model parameters are optimized by minimizing the mean squared error (MSE) between predicted outputs $\hat{y}$ and ground-truth values $y$.

\section{Experimental Results}
\subsection{Experimental Setup}
\begin{figure*}[t]
\centering
\includegraphics[width=\textwidth]{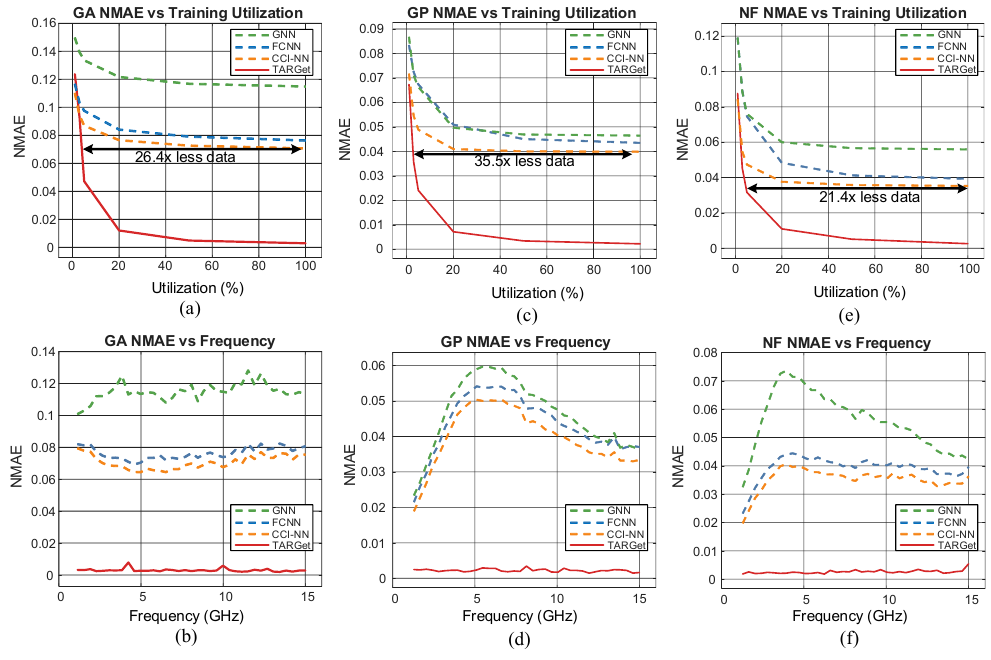}
\caption{Model performance error (NMAE) for the LNA architecture across different specifications. (a,c,e) NMAE vs. training data utilization for GA, GP, and NF, respectively. (b,d,f) NMAE vs. frequency for GA, GP, and NF, respectively.}
\Description{Six line plots comparing TARGet with GNN, FCNN, and CCI-NN on the LNA dataset. The top row plots NMAE against training utilization for GA, GP, and NF, with arrows indicating large data reduction factors. The bottom row plots NMAE against frequency for the same specifications. In all panels, the TARGet curve stays below the baseline curves.}
\label{fig:lna_error}
\end{figure*}

\begin{table}[t]
\centering
\caption{Performance comparison on the LNA architecture (lower is better).}
\label{tab:main_results}
\begin{tabular}{llcccc}
\toprule
Spec & Metric & FCNN & CCI-NN & GNN & TARGet \\
\midrule
\multirow{3}{*}{GA} & MAE   & 0.3417 & 0.3176 & 0.5019 & \textbf{0.0132} \\
 & RMSE  & 0.4968 & 0.4655 & 0.6682 & \textbf{0.0198} \\
 & NMAE  & 0.0784 & 0.0730 & 0.1145 & \textbf{0.0030} \\
\midrule
\multirow{3}{*}{GP} & MAE   & 0.2206 & 0.2015 & 0.2354 & \textbf{0.0126} \\
 & RMSE  & 0.2429 & 0.2131 & 0.2510 & \textbf{0.0240} \\
 & NMAE  & 0.0414 & 0.0378 & 0.0442 & \textbf{0.0024} \\
\midrule
\multirow{3}{*}{NF} & MAE   & 0.1499 & 0.1339 & 0.2127 & \textbf{0.0107} \\
 & RMSE  & 0.1867 & 0.1614 & 0.2375 & \textbf{0.0177} \\
 & NMAE  & 0.0376 & 0.0336 & 0.0532 & \textbf{0.0027} \\
\bottomrule
\end{tabular}
\end{table}

\begin{table}[t]
\centering
\caption{Tolerance-based evaluation for the LNA architecture (higher is better).}
\label{tab:tolerance}
\begin{tabular}{lcccc}
\toprule
Model & $\leq$0.01 & $\leq$0.02 & $\leq$0.05 & $\leq$0.10 \\
\midrule
FCNN~\cite{Fukuda2017}   & 0.0987 & 0.1988 & 0.4901 & 0.7664 \\
CCI-NN~\cite{Hassanpourghadi2021} & 0.0940 & 0.1978 & 0.5350 & 0.7997 \\
GNN~\cite{zhang2019circuit}    & 0.0526 & 0.1075 & 0.2997 & 0.5984 \\
TARGet & \textbf{0.9582} & \textbf{0.9936} & \textbf{0.9998} & \textbf{1.0000} \\
\bottomrule
\end{tabular}
\end{table}

\begin{figure}[htbp]
\centering
\includegraphics[width=\columnwidth]{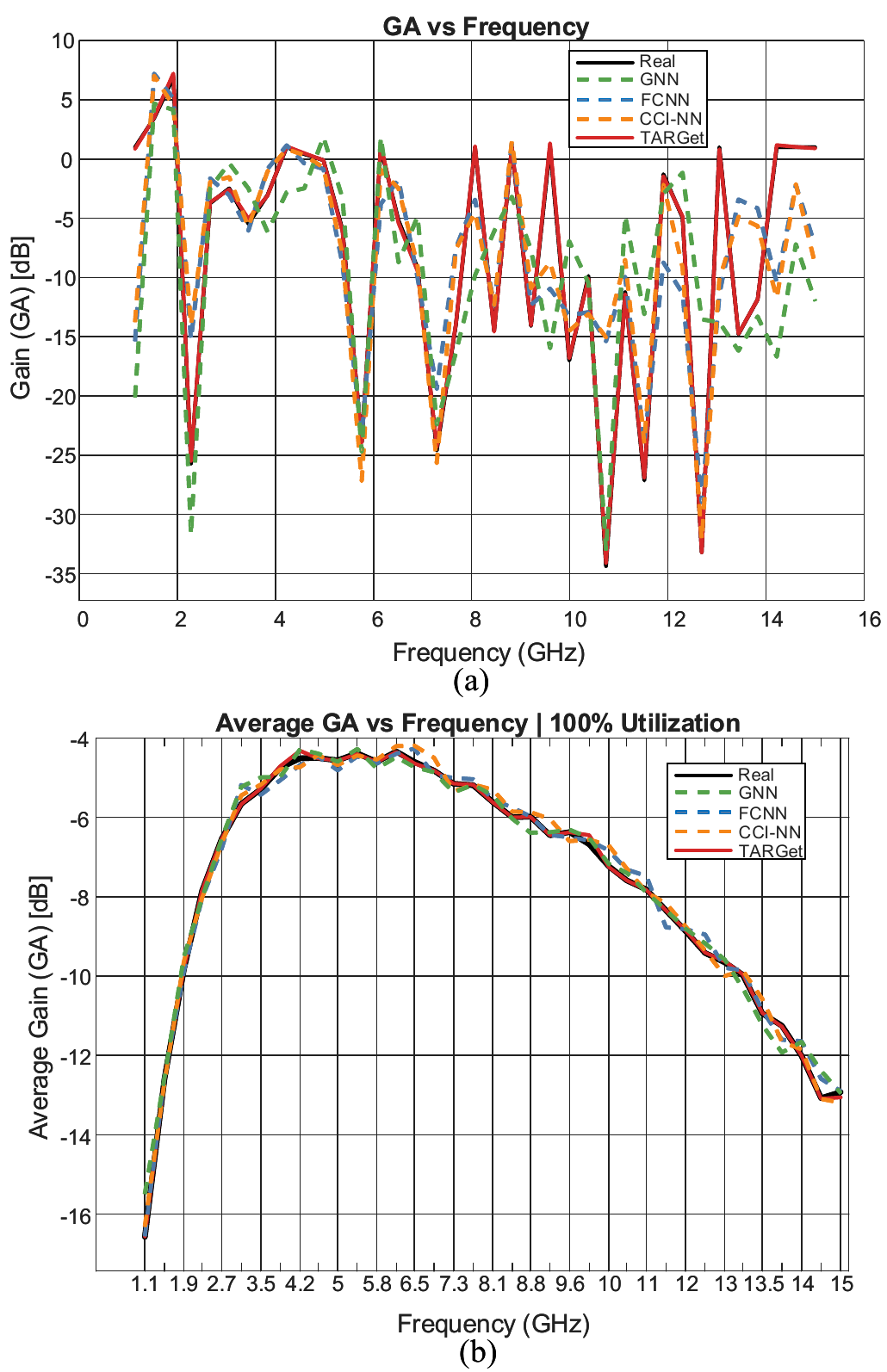}
\caption{Comparison of predicted and ground-truth gain (GA) across frequency for the LNA architecture at full training data utilization. (a) Representative sample-level predictions. (b) Average GA across all samples, showing close agreement between model predictions and ground truth.}
\Description{Two plots of available gain versus frequency. The top plot compares real and predicted GA curves for a representative sample using TARGet, GNN, FCNN, and CCI-NN. The bottom plot compares the average GA over all samples. TARGet closely follows the real curve, especially in the averaged response.}
\label{fig:ga}
\end{figure}

We evaluate TARGet across multiple RF circuit datasets, including phase-shifter and low-noise amplifier (LNA) architectures, each comprising multiple topologies. The supported high-level topologies and corresponding sub-circuit building blocks are illustrated in Fig.~\ref{fig:database}. The LNA dataset includes 98 distinct topologies constructed from combinations of sub-circuit modules, as illustrated in Fig.~\ref{fig:database}(a)--(b). Similarly, the phase-shifter dataset consists of 36 different topologies, shown in Fig.~\ref{fig:database}(c)--(d). The statistical properties of the target specifications are summarized in Tables~\ref{tab:spec_stat_phase_shifter} and ~\ref{tab:spec_stat_lna}.

The datasets consist of design parameters and corresponding performance specifications across a wide range of operating frequencies. Each circuit is represented using sub-circuit S-parameters in Cartesian form, auxiliary features, and topology information. The model predicts performance metrics such as gain and reflection coefficients. The datasets are generated using a 55 nm BiCMOS PDK, where inductors and capacitors are modeled using realistic device characteristics, including quality factor and self-resonance effects. Circuit simulations are performed across a wide frequency range from 1~Hz to 15~GHz.

The datasets are preprocessed by removing invalid samples, applying outlier clipping, and standardizing input and output features. Topology information is encoded using one-hot representations when applicable. To evaluate data efficiency, models are trained at different utilization levels, where only a fraction of the available training data is used. Performance is evaluated in terms of mean absolute error (MAE), root mean squared error (RMSE), and normalized mean absolute error (NMAE), which enables fair comparison across different circuit architectures. NMAE is computed as the mean absolute error normalized by the output range for the corresponding specification and frequency setting.

To evaluate topology-level transfer, we also introduce a stricter zero-shot held-out split for the LNA dataset in which all samples containing a selected matching-network topology are excluded from training, validation, and standard testing and evaluated only after training. The original LNA experiments use 40 frequency points, while the held-out evaluation uses 10 uniformly selected frequencies.

All models are implemented using TensorFlow/Keras, with graph-based components built using the Spektral library. Training is performed using the Adam optimizer with mean squared error (MSE) loss and early stopping based on validation loss. Experiments were conducted on a high-performance computing cluster equipped with NVIDIA A100 (40 GB) GPUs.
To quantify the overhead of the fused architecture, we measure training time and single-sample inference latency averaged across held-out runs. TARGet incurs a longer one-time training duration (130.7 s versus 30-61 s for the baselines), while its 62.3 ms single-sample inference latency remains comparable to the 48.5-55.7 ms baseline range.

\subsection{Models and Baselines}
We compare TARGet against several baseline models that capture different aspects of circuit modeling. These include a fully connected neural network (FCNN), a circuit connectivity-inspired neural network (CCI-NN), and a graph neural network (GNN). The architectural details of all models are summarized in Table~\ref{tab:architecture_all}.

These models differ in how they represent circuit structure, ranging from flat feature representations (\textit{e.g.}, FCNN) to topology-aware modeling (\textit{e.g.}, GNN), while TARGet combines both sub-circuit and topology-level information within a unified framework. A high-level comparison of these models is provided in Table~\ref{tab:model_comparison}.

\subsection{Results on LNA Architecture}
\subsubsection{\textbf{Data Efficiency and Quantitative Results}}
We analyze model performance on the LNA architecture across varying levels of training data utilization. Each model is trained on different fractions of the training set and evaluated on a fixed test set. Utilization denotes the fraction of data used, enabling evaluation under data-scarce conditions.

TARGet consistently achieves lower prediction error across all specifications (GA, GP, and NF) and utilization levels, as shown in Fig.~\ref{fig:lna_error}. For example, in the GA specification, TARGet reduces NMAE from 0.0730 (CCI-NN) to 0.0030, corresponding to over 95\% improvement (approximately 24x lower error), with particularly large gains in low-data regimes.

Moreover, TARGet exhibits consistently low prediction error across the entire frequency range, indicating robust generalization across operating conditions. Notably, TARGet achieves performance comparable to the best baseline models while using only 2.8\%-4.7\% of the training data (corresponding to a 21.4x-35.5x reduction in training data utilization). Table~\ref{tab:main_results} summarizes the overall prediction performance across different specifications and error metrics.

To further illustrate prediction quality, Fig.~\ref{fig:ga} compares predicted and ground-truth gain (GA) across frequency. TARGet closely follows the ground-truth trends for both individual samples and averaged responses, demonstrating accurate modeling of frequency-dependent behavior.

\begin{table}[t]
\centering
\caption{Zero-shot transfer NMAE results on the held-out LNA topology using 10 uniformly selected frequencies at 100\% utilization (lower is better).}
\label{tab:zero_shot}
\begin{tabular*}{0.7\columnwidth}{@{\extracolsep{\fill}} lccc}
\toprule
Model & GA & GP & NF \\
\midrule
FCNN & 0.0856 & 0.0604 & 0.0649 \\
CCI-NN & 0.0833 & 0.0593 & 0.0671 \\
GNN & 0.1185 & 0.0656 & 0.0837 \\
TARGet & \textbf{0.0457} & \textbf{0.0482} & \textbf{0.0499} \\
\bottomrule
\end{tabular*}
\end{table}

\subsubsection{\textbf{Tolerance-Based Evaluation}}
\begin{figure}[t]
\centering
\includegraphics[width=0.8\columnwidth]{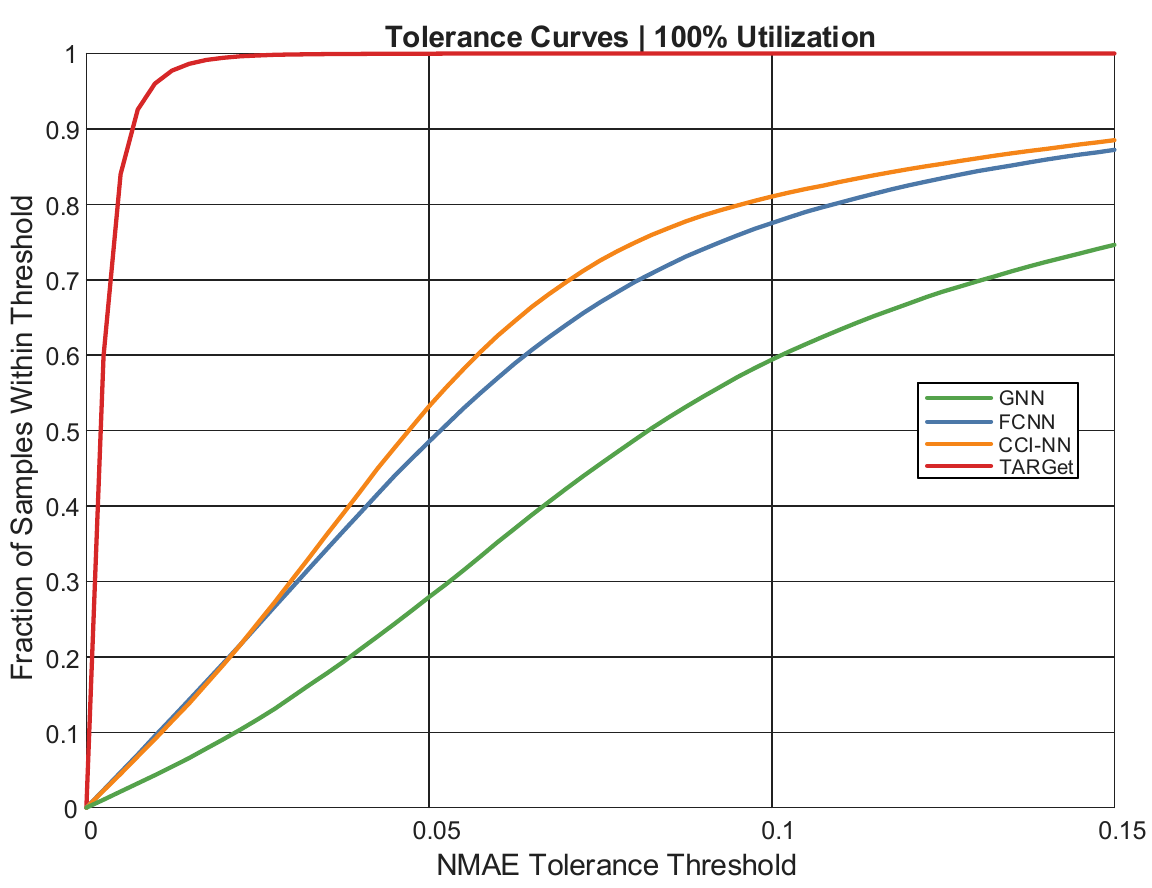}
\caption{Tolerance curves for the LNA architecture at full training data utilization. The curves show the fraction of predictions whose normalized error falls below a given threshold.}
\Description{Cumulative tolerance plot with NMAE tolerance threshold on the horizontal axis and fraction of samples within the threshold on the vertical axis. TARGet rises sharply and reaches nearly all samples at a low threshold, while FCNN, CCI-NN, and GNN increase more gradually.}
\label{fig:tolerance_curve}
\end{figure}

For the LNA architecture, we further evaluate prediction robustness under varying error tolerance thresholds. Specifically, we measure the fraction of predictions whose normalized absolute error falls below a given threshold. As shown in Fig.~\ref{fig:tolerance_curve}, TARGet significantly outperforms baseline models across all thresholds, achieving near-perfect accuracy even under strict error constraints.

\begin{figure}[htbp]
\centering
\includegraphics[width=\columnwidth]{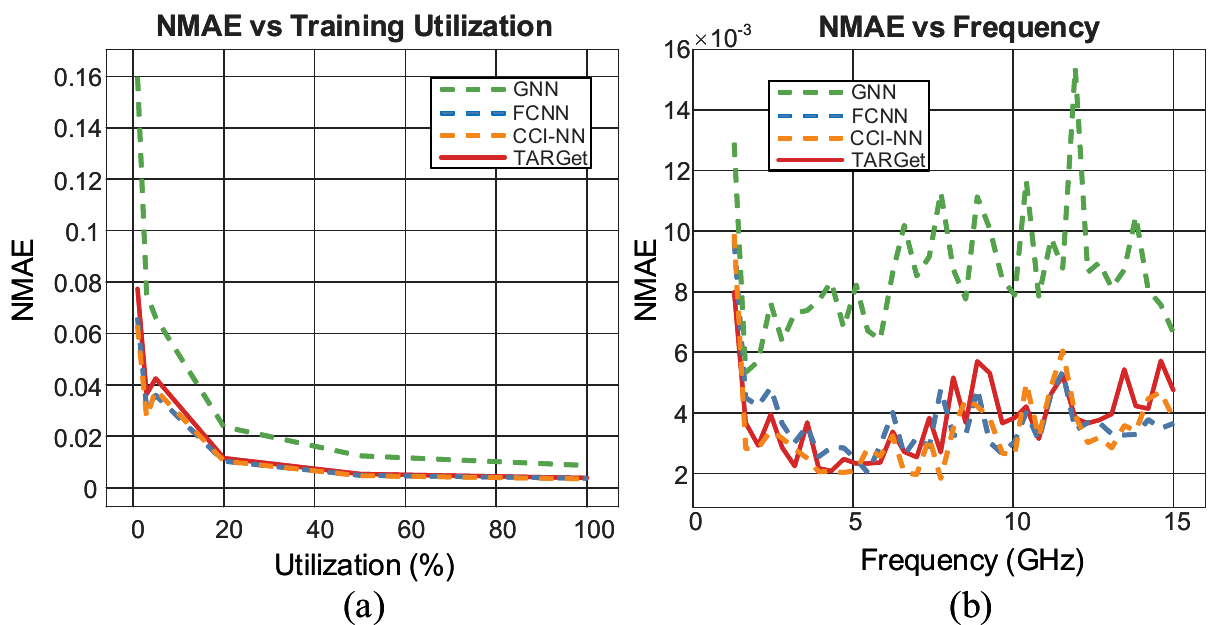}
\caption{Model performance error (NMAE) for the phase-shifter architecture for the input return loss specification. (a) NMAE as a function of training data utilization, illustrating model performance under data-scarce conditions. (b) NMAE across frequency, showing prediction consistency over the operating range.}
\Description{Two line plots for the phase-shifter dataset. The left plot shows NMAE versus training utilization, where all methods improve as utilization increases and TARGet, FCNN, and CCI-NN are close at high utilization. The right plot shows NMAE versus frequency, where GNN has higher error and the other three methods remain lower and close to each other.}
\label{fig:ps_error}
\end{figure}

As shown in Table~\ref{tab:tolerance}, TARGet achieves significantly higher accuracy across all tolerance levels. In particular, over 95\% of predictions fall within a 0.01 error threshold, whereas baseline models remain below 20\% at the same threshold, corresponding to nearly 9.7x higher accuracy, highlighting TARGet's ability to produce highly precise predictions under strict error constraints.

\subsection{Zero-Shot Topology Evaluation}
To evaluate topology-level transfer, we perform a held-out-topology LNA evaluation. In this split, an entire matching-network topology is removed from the main train/validation/test pool wherever it appears on either the input or output matching side and is evaluated only as zero-shot data, as shown in Fig.~\ref{fig:database}(b). This creates more than 750,000 held-out zero-shot samples, while the remaining LNA data contains over 4.8 million training, 1.2 million validation, and 1.5 million test samples.

As shown in Table~\ref{tab:zero_shot}, TARGet achieves the lowest held-out NMAE for all three LNA specifications. Relative to the best baseline for each metric, TARGet reduces zero-shot NMAE by up to \textbf{45.1\%}.

\subsection{Results on Phase-Shifter Architecture}
As shown in Fig.~\ref{fig:ps_error}, compared to the LNA case, all models achieve similar performance on the phase-shifter dataset, with only minor differences between methods. Notably, simpler models such as CCI-NN slightly outperform TARGet in this case. This behavior can be attributed to the relatively simple structure of phase shifter circuits, where the underlying relationships can be learned effectively even without explicit topology modeling, making the advantage of topology-aware approaches less pronounced. Nevertheless, TARGet remains competitive, indicating that the proposed framework does not degrade performance in simpler scenarios while providing substantial gains for more complex architectures such as LNA.

\section{Conclusion}
In this work, we proposed TARGet, an open-source topology-aware RF circuit modeling framework that jointly captures sub-circuit interactions and high-level circuit structure. By integrating sub-circuit connectivity-aware modeling with graph-based topology encoding, TARGet effectively preserves structural information that is often ignored in conventional learning approaches. Evaluating TARGet on multiple RF circuit architectures, including LNAs and phase shifters, demonstrates that TARGet achieves a normalized error below 1\%. Moreover, compared to SOTA methods, while maintaining the same error, TARGet reduces the required training data by up to 35.5x. Furthermore, TARGet achieves 9.7x higher prediction accuracy at a strict 1\% error threshold compared to SOTA models. In particular, TARGet maintains strong performance under limited training data and exhibits robust generalization across frequency, while reducing NMAE by up to 45\% relative to the best baseline in the zero-shot evaluation.

\bibliographystyle{ACM-Reference-Format}
\bibliography{Bibliography}

\end{document}